\documentclass[letterpaper, 10pt, conference]{ieeeconf}  % Comment this line out if you need a4paper

\IEEEoverridecommandlockouts                              % This command is only needed if 
\usepackage[utf8]{inputenc}
\usepackage[english]{babel}
\usepackage{amsmath}
\usepackage{amssymb}
\usepackage{graphicx}
\usepackage{subcaption}
\usepackage{hyperref}
\usepackage{url}
\usepackage{xcolor}

\title{\LARGE \bf Fast and Robust Initialization for Visual-Inertial SLAM}
\author{Carlos Campos, José M.M. Montiel and Juan D. Tardós
\thanks{This work was supported in part by the Spanish government under grants DPI2015-67275 and DPI2017-91104-EXP, the Aragón government under grant DGA\_T45-17R, and by Huawei under grant HF2017040003.}% <-this % stops a space
\thanks{The authors are with Instituto de Investigación en ingeniería de Aragón (I3A), Universidad de Zaragoza, Spain   {\tt\small campos@unizar.es; josemari@unizar.es; tardos@unizar.es}}%
}

\begin{document}

\thispagestyle{empty}
\newpage
\onecolumn

\begin{center}
This paper has been published in 2019 International Conference on Robotics and Automation (ICRA).
\vspace{0.75cm}

DOI: \href{https://doi.org/10.1109/ICRA.2019.8793718}{\textcolor{blue}{10.1109/ICRA.2019.8793718}}\\
IEEE Xplore: \href{https://ieeexplore.ieee.org/document/8793718}{\textcolor{blue}{https://ieeexplore.ieee.org/document/8793718}}
\vspace{1.25cm}
\end{center}

©2019 IEEE. Personal use of this material is permitted. Permission from IEEE must be obtained for all other uses, in any current or future media, including reprinting/republishing this material for advertising or promotional purposes, creating new collective works, for resale or redistribution to servers or lists, or reuse of any copyrighted component of this work in other works.

\twocolumn

\maketitle
\thispagestyle{empty}
\pagestyle{empty}

\begin{abstract}
Visual-inertial SLAM (VI-SLAM) requires a good initial estimation of the initial velocity, orientation with respect to gravity and gyroscope and accelerometer biases. In this paper we build on the initialization method proposed by Martinelli \cite{martinelli2014closed} and extended by Kaiser et al. \cite{kaiser2017simultaneous}, modifying it to be more general and efficient. We improve accuracy with several rounds of visual-inertial bundle adjustment, and robustify the method with novel observability and consensus tests, that discard erroneous solutions. Our results on the EuRoC dataset show that, while the original method produces scale errors up to 156\%,  our method is able to consistently initialize in less than two seconds with scale errors around 5\%, which can be further reduced to less than 1\% performing visual-inertial bundle adjustment after ten seconds.

\end{abstract}

%%%%%%%%%%%%%%%%%%%%%%%%%%%%%%%%%%%%%%%%%%%%%%%
\section{INTRODUCTION}

Visual-Inertial SLAM stands for those techniques able to build a map and simultaneously locate an agent inside the map, using as input a camera and an Inertial Measurement Unit (IMU)  \cite{li2013high} \cite{leutenegger2015keyframe} \cite{mur2017visual}. Both sensors require to be initialized or calibrated before using them. While camera is calibrated just once as it does not evolve with time, IMU has to be initialized before every use. Inertial initialization aims to compute values for the initial velocity, gravity direction and gyroscope and accelerometer biases that can be used as initial seed for the Visual-Inertial SLAM estimation process, guaranteeing its convergence. Once these values are found, inertial measurements can be used to improve tracking, making it more robust, as well as to find the true scale of the map, which cannot be obtained with pure monocular systems.

Some techniques initialize first the monocular SLAM part, building a map up-to-scale, and then, enough time later for having good observability of all variables, compute scale factor, gravity direction and IMU biases. This technique, first proposed by Mur-Artal and Tardós \cite{mur2017visual} and latter adapted in \cite{qin2017robust}, gives very accurate results. In particular, the true scale of the environment can be estimated with an error around 1\%. However, this approach has two main drawbacks:
\begin{itemize}
	\item Dependency on visual initialization. It requires to have an initial map up-to-scale with enough map-points. If the visual part is not able to initialize, or if it gets lost within a short period of time, the inertial system will not initialize.
	\item Initialization takes too long. The method requires to run monocular SLAM for around fifteen seconds to safely find the correct inertial parameters for the visual-inertial bundle adjustment (BA). This time is too long for many applications.
\end{itemize}

Martinelli \cite{martinelli2014closed} proposed a very interesting  closed form method that only requires to track a few points, and solves jointly for the camera motion, the environment structure, and the inertial parameters. The paper included a theoretical analysis of the conditions for the problem to be solvable and presented simulated results. The recent work of Kaiser et al.  \cite{kaiser2017simultaneous} extends the method to also estimate gyroscope bias, and presents results with real data from a drone, showing that it is possible to initialize in less than 3 seconds, with relative errors around 15\% on speed and gravity. This initialization method has several limitations: 
\begin{itemize}
	\item It assumes that all features are tracked in all frames, and that all tracks provided are correct. In case of spurious tracks, it can provide arbitrarily bad solutions.
	\item The initialization accuracy is low, compared to \cite{mur2017visual}. To improve it, a lot of tracks and frames are needed, increasing  its computational cost, and making it unfeasible in real time. 
	\item With noisy sensors, trajectories that are close to the unsolvable cases analyzed in \cite{martinelli2014closed}  give weak observability of some of the variables. The method lacks robust criteria to decide when an initialization is accurate enough.
\end{itemize}

In this paper we build on the \textit{Martinelli-Kaiser solution} \cite{martinelli2014closed} \cite{kaiser2017simultaneous}  (or simply \textit{MK-solution}), modifying it to be more general, efficient, robust and precise. The main novelties of our initialization algorithm are:

\begin{enumerate}
	\item \textbf{Generality}: we generalize the method to use partial tracks and to take into account the camera-IMU relative pose.
	\item \textbf{Efficiency}:  we reduce the running time by using a fixed number of $m$ features and $n$ keyframes carefully chosen, and adopting the preintegration method proposed in \cite{lupton2012visual} \cite{forster2015imu}.
	\item \textbf{Observability test}: after \textit{MK-solution}, we perform visual-inertial BA with the $m$ points and $n$ keyframes, and apply a novel observability test to check the accuracy of the solution. If the test fails, the initialization is discarded.
	\item \textbf{Consensus test}: we try to initialize additional tracked features by triangulating the 3D point and checking the reprojection error.  If enough consensus is found, we perform a second visual-inertial BA with all points and keyframes, and initialize the SLAM map with them. Otherwise, the initialization is discarded.
\end{enumerate}

We present exhaustive initialization tests on the EuRoC dataset  \cite{burri2016euroc} showing that the method is able to consistently initialize in less than 2s with a scale error of 5\%. This error  converges to 1\% performing visual-inertial BA after 10s, when the trajectory is long enough to give good observability of all variables.

\section {Initial Solution}

\subsection{Feature extraction and tracking}

We use the multiscale ORB extractor \cite{rublee2011orb} to find $M$ uniformly distributed points to be tracked. As points with high resolution are preferable, we only extract ORB points at the three finest image scales, using a scale factor of 1.2. We have found that the extraction of FAST features and matching with ORB descriptors is not stable enough to obtain long tracks, as most of the features were lost in some of the frames. Hence, to solve this problem, we have performed feature tracking using the pyramidal implementation of the Lucas-Kanade algorithm \cite{lucas1981iterative} available in OpenCV.

However, simple tracking is not enough since there could be some tracks which are not correct, particularly in areas of repetitive or low texture. To attack this problem, the geometric consistency of these tracks is checked by finding a fundamental matrix using RANSAC. The combination of ORB keypoints, Lucas-Kanade tracking (KLT), and the fundamental matrix check, leads to long tracks for a high number of features. When some track is lost or rejected by the RANSAC algorithm, new ORB points are extracted and start to be tracked, in order to keep a constant number of $M$ tracks. Each time we detect that there are at least $m$ tracked points with a track length of at least $l$ pixels, we launch our initialization method. This first test to decide whether to attempt initialization or not is called \textit{track-length test}.

\subsection{Modified Martinelli-Kaiser solution}
\label{sec:InitialSolution}
In this part we extend the method proposed in \cite{martinelli2014closed} \cite{kaiser2017simultaneous} to be able to deal with features not seen by all cameras, and computing terms in an efficient way using inertial pre-integration proposed by \cite{lupton2012visual} \cite{forster2015imu}.
Let $m$ be the number of tracked features used for initialization, placed at $(\textbf{x}_1 \dots \textbf{x}_m)$ in the global reference frame, and $\mathcal{C}=\{\mathcal{C}_1 \dots \mathcal{C}_n\}$ the set of $n$ cameras indexes used for initialization, also referred to as keyframes, which are chosen to be uniformly distributed along time. Let's also call $\mathcal{C}^i = \{\mathcal{C}_{1_i} \dots \mathcal{C}_{n_i}\}$ the set of $n_i$ cameras indexes seeing feature $i$-th. Here we remark that, in contrast to \cite{martinelli2014closed}, in our formulation not all features have to be observed by all cameras, and the transformation from camera to body (IMU), obtained from calibration, is taken into account. Without loss of generality we use as global reference the first body pose used for initialization.

\begin{figure}
  \centering
  \includegraphics[width=0.50\textwidth]{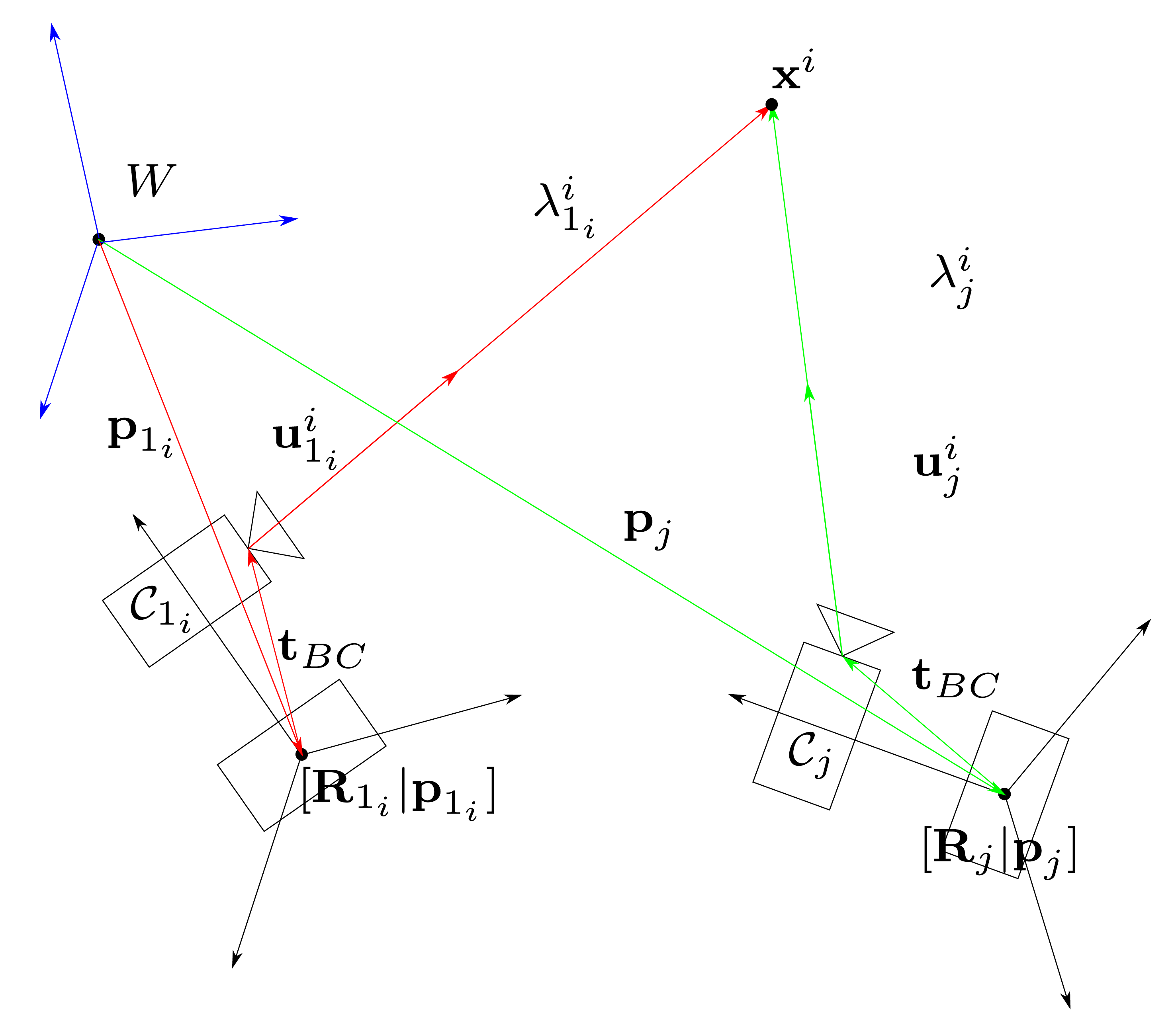}
  \caption{Relationships between two Body (IMU) and Camera poses observing the same feature.}
  \label{fig:scheme2}
\end{figure}

From figure \ref{fig:scheme2}, we can write the set of equalities:

\begin{equation} \label{eq:init1}
\begin{split}
\textbf{p}_{1_i} + \textbf{R}_{1_i}\textbf{t}_{BC} + \lambda_{1_i}^i \textbf{u}_{1_i}^i  = \textbf{p}_{j} + \textbf{R}_{j} \textbf{t}_{BC} + \lambda_j^i \textbf{u}_j^i  \\ i = 1 \dots m, \quad \forall j \in \mathcal{C}^i \setminus 1_i
\end{split}
\end{equation}

\noindent where:
\begin{itemize}
\item $\textbf{p}_j \in  \mathbb{R}^3$: position of the $j$-th body in the global reference frame.
\item $\textbf{R}_j \in \text{SO}(3)$: rotation matrix from $j$-th body to global reference frame.
\item $\lambda_j^i$: distance between $i$-th feature and $j$-th camera.
\item $\textbf{u}_j^i$: unitary vector from $j$-th camera to $i$-th feature in the global reference frame. It can be computed as  $\textbf{u}_j^i = \textbf{R}_{j} \textbf{R}_{BC}\,{}_{c_j}\textbf{u}_j^i$, being ${}_{c_j}\textbf{u}_j^i$ the unitary vector in the $j$-th camera reference frame for the $i$-th feature, which is directly obtained from the tracked images.
\item $[\textbf{R}_{BC} \vert \textbf{t}_{BC}]$: transformation from Camera to Body (IMU).
\end{itemize}

$\textbf{R}_{j}$ and $\textbf{p}_j$ can be computed by integrating the  inertial measurements. Considering constant biases during the initialization time, it leads to \cite{forster2015imu}:

\begin{equation}
\label{eq:rotation}
\textbf{R}_{1,j}(\textbf{b}^g)  = \prod_{k=1}^{t_j-1} \text{Exp} (( \pmb{\omega}_k^m - \textbf{b}^{g}) \Delta t) 
\end{equation}
\begin{equation}
\label{eq:velocity}
\textbf{v}_{j}(\textbf{b}) = \textbf{v}_{1} + \textbf{g} \Delta t_{1,j} + \sum_{k=1}^{t_j-1} \textbf{R}_{1,k} (\textbf{a}_k^m - \textbf{b}^{a}) \Delta t
\end{equation}
\begin{equation}
\label{eq:position}
\begin{split}
\textbf{p}_{j}(\textbf{b}) = \textbf{p}_{1} + \sum_{k=1}^{t_j-1} \left( \textbf{v}_{k} \Delta t + \frac{1}{2} \textbf{g} \Delta t^2 + \frac{1}{2} \textbf{R}_{1,k} ( \textbf{a}_k^m - \textbf{b}^a) \Delta t^2  \right)
\end{split}
\end{equation}

\noindent where:
\begin{itemize}
\item $\text{Exp}$ stands for the exponential map $\text{Exp}: \mathbb{R}^3 \rightarrow \text{SO}(3)$
\item $\textbf{a}_k^m$ and $\pmb{\omega}_k^m$ are $k$-th acceleration and angular velocity IMU measurement
\item $\textbf{b} = (\textbf{b}^a, \textbf{b}^g)$ are their corresponding accelerometer and gyro biases
\item $\textbf{g}$ stands for gravity in the global frame
\item $\textbf{v}_j$ is the velocity at the $j$-th body.
\item $\Delta t_{i,j}$ denotes time difference between $i$-th and $j$-th poses.
\end{itemize}

However, the above expressions have an important drawback: each time that $\textbf{b}^g$ or $\textbf{b}^a$ are modified, all the IMU integration requires to be recomputed, which is very time consuming. To solve this problem we have adopted the linear preintegration correction from \cite{lupton2012visual} \cite{forster2015imu},  splitting these expressions in bias dependent and non-dependent terms, using delta terms $\Delta \textbf{R}_{1,j}$, $\Delta \textbf{v}_{1,j}$, $\Delta \textbf{p}_{1,j}$, which are directly computed from IMU measurements and which are defined as follows:

\begin{equation} \label{eq:DeltaR}
\Delta \textbf{R}_{1,j} \triangleq \textbf{R}_{1}^T \textbf{R}_{j}
\end{equation}
\begin{equation} \label{eq:DeltaV}
\Delta \textbf{v}_{1,j} \triangleq \textbf{R}_{1}^T (\textbf{v}_j-\textbf{v}_1-\textbf{g} \Delta t_{1,j})
\end{equation}
\begin{equation} \label{eq:DeltaP}
\Delta \textbf{p}_{1,j} \triangleq \textbf{R}_{1}^T (\textbf{p}_j-\textbf{p}_1-\textbf{v}_1 \Delta t_{1,j}- \frac{1}{2} \textbf{g} \Delta t_{1,j}^2 )
\end{equation}

If during intermediate steps $\textbf{b}^g$ changes more than 0.2 rad/sec from the value used for preintegration, the preintegration is recomputed with this new bias, otherwise, delta terms are directly updated using their Jacobians w.r.t. biases ($\frac{\partial \Delta \textbf{R}_{1,j}}{\partial \textbf{b}^g}$, $\frac{\partial \Delta \textbf{v}_{1,j}}{\partial \textbf{b}^g}$, $\frac{\partial \Delta \textbf{v}_{1,j}}{\partial \textbf{b}^a}$, $\frac{\partial \Delta \textbf{p}_{1,j}}{\partial \textbf{b}^g}$, $\frac{\partial \Delta \textbf{p}_{1,j}}{\partial \textbf{b}^a}$). In this way, we relinearize each time we get too far from the linearization point. These Jacobians can be found in \cite{forster2015imu}. We rewrite eq. (\ref{eq:init1}) using expressions (\ref{eq:DeltaR}), (\ref{eq:DeltaV}) and (\ref{eq:DeltaP}):

\begin{equation} \label{eq:singularSys1}
\begin{split}
\lambda_{1_i}^i \textbf{u}_{1_i}^i - \lambda_j^i \textbf{u}_j^i  -  \textbf{v}_{1} \Delta t_{1_i,j} - \textbf{g} \left( \frac{\Delta t_{1,j}^2-\Delta t_{1,1_i}^2}{2} \right) = \\ \Delta \textbf{p}_{1,j} - \Delta \textbf{p}_{1,1_i} + (\Delta \textbf{R}_{1,j} -\Delta \textbf{R}_{1,1_i})\textbf{t}_{BC} \\
\forall i = 1 \dots m, \quad \forall j \in \mathcal{C}^i \setminus 1_i
\end{split}
\end{equation}

%$$ \textbf{s}_j (\textbf{b}^g, \textbf{b}^a)$$

Now, we can add the gravity magnitude information. Instead of adding it as a constraint of the gravity magnitude as done in \cite{kaiser2017simultaneous}, we prefer to model the gravity by mean of a rotation matrix parametrized by only two angles $(\alpha, \beta)$ (rotation around $z$-axis has no effect) and the vector $\textbf{g}_{I} = (0,0,-g)$, thus we remain in an unconstrained problem:

\begin{equation} \label{eq:sys}
\textbf{g}  = \text{Exp}(\alpha,\beta,0) \textbf{g}_{I}
\end{equation}

Equation (\ref{eq:singularSys1}) becomes:

\begin{equation} \label{eq:singularSys2}
\begin{split}
\lambda_{1_i}^i \textbf{u}_{1_i}^i - \lambda_j^i  \textbf{u}_j^i -  \textbf{v}_{1} \Delta t_{1_i,j} = \textbf{s}_{1_i,j} (\textbf{b}^g, \textbf{b}^a, \alpha, \beta)
\end{split}
\end{equation}

\noindent where:

\begin{equation} \label{eq:s_vector}
\begin{split}
\textbf{s}_{1_i,j} (\textbf{b}^g, \textbf{b}^a, \alpha, \beta) = \text{Exp}(\alpha,\beta,0) \textbf{g}_{I} \left( \frac{\Delta t_{1,j}^2-\Delta t_{1,1_i}^2}{2} \right)  + \\
\Delta \textbf{p}_{1,j} - \Delta \textbf{p}_{1,1_i} + (\Delta \textbf{R}_{1,j}-\Delta \textbf{R}_{1,1_i})\textbf{t}_{BC}
\end{split}
\end{equation}

Now, each time biases are updated, we do not need to preintegrate again all the measurements, but only to update them by means of Jacobians. Neglecting accelerometer bias as in \cite{kaiser2017simultaneous}, the only unknowns are $\lambda_j^i \ (\forall i=1 \dots m, \, j \in \mathcal{C}^i \setminus 1_i)$, $\textbf{v}_1$, $\alpha$, $\beta$ and $\textbf{b}^g$. Stacking equations for all possible values of $i$ and $j$ we build an overdetermined sparse linear system, with only three non-zero elements per row, such as:

\begin{equation} \label{eq:sys}
\textbf{A}(\textbf{b}^g)\textbf{x}=\textbf{s}(\textbf{b}^g, \alpha, \beta)
\end{equation}

where $\textbf{x} = (\textbf{v}_1, \{ \lambda_j^i \})$ is the unknown vector with linear dependence. To jointly find linear and no-linear dependent parameters, we solve the next unconstrained minimization problem:

\begin{equation} \label{eq:optim}
(\textbf{b}^g, \alpha, \beta) \triangleq \underset{\textbf{b}^g, \alpha, \beta}{\arg \min} \left( \underset{\textbf{x}}{\min} \left\lVert \textbf{A}(\textbf{b}^g)\textbf{x} - \textbf{s}(\textbf{b}^g, \alpha, \beta) \right\lVert^2_2 \right)
\end{equation} 

Cost function $c(\textbf{b}^g, \alpha, \beta) = \underset{\textbf{x}}{\min} \left\lVert \textbf{A}(\textbf{b}^g)\textbf{x} - \textbf{s}(\textbf{b}^g, \alpha, \beta) \right\lVert^2_2$ is evaluated for each $(\textbf{b}^g, \alpha, \beta)$ using the following scheme:
\begin{enumerate}
\item \textbf{Update} $\Delta \textbf{R}_{1,j}$ \textbf{and} $\Delta \textbf{p}_{1,j}$: Using \cite{forster2015imu}, we don't need to reintegrate all IMU measurements each time that $\textbf{b}^g$ changes. We simply update delta terms using their Jacobians w.r.t. bias. That supposes an important computational saving.
\item \textbf{Compute} $\textbf{A}(\textbf{b}^g)$ \textbf{and} $\textbf{s}(\textbf{b}^g, \alpha, \beta)$ and build the linear system.
\item \textbf{Solve} $\textbf{x} = \textbf{A}(\textbf{b}^g)$\textbackslash$\textbf{s}(\textbf{b}^g, \alpha, \beta)$ using conjugate gradient, which is suitable for sparse systems.
\item \textbf{Compute} $c(\textbf{b}^g, \alpha, \beta) = \left\lVert \textbf{A}(\textbf{b}^g)\textbf{x} - \textbf{s}(\textbf{b}^g, \alpha, \beta) \right\lVert^2_2$.
\end{enumerate}

The computational cost of evaluating $c(\textbf{b}^g, \alpha, \beta)$ comes, first, from solving a sparse system with no more than $3+n \times m$ unknowns and $3 \times (n-1) \times m$ equations and, second, from integrating inertial measurement along the initialization time. However, using formulation from \cite{forster2015imu}, we can avoid reintegration, integrating IMU measures only once, and updating preintegrated terms by means of a linear approximation.

To optimize  $c(\textbf{b}^g, \alpha, \beta)$ and find the correct gyro bias and gravity direction we use Levenberg–Marquardt algorithm, where Jacobians of the cost function are computed numerically. As result, not only IMU initialization parameters are found ($\textbf{g}$, $\textbf{b}^g$ and $\textbf{v}_1$) but also the position of tracked points ($\lambda_j^i \textbf{u}_j^i$). We highlight that not all $M$ tracked features have been used during this initialization, but only a small set of $m$ features, aiming to reduce computational complexity. However, the solutions found after this step are not accurate enough to launch the system, and further intermediate stages are required.

\section{Improved Solution}
\subsection{First BA and observability test}
 \label{sec:BA1_Obs}
 
\begin{figure}
  \centering
  \includegraphics[width=0.49\textwidth]{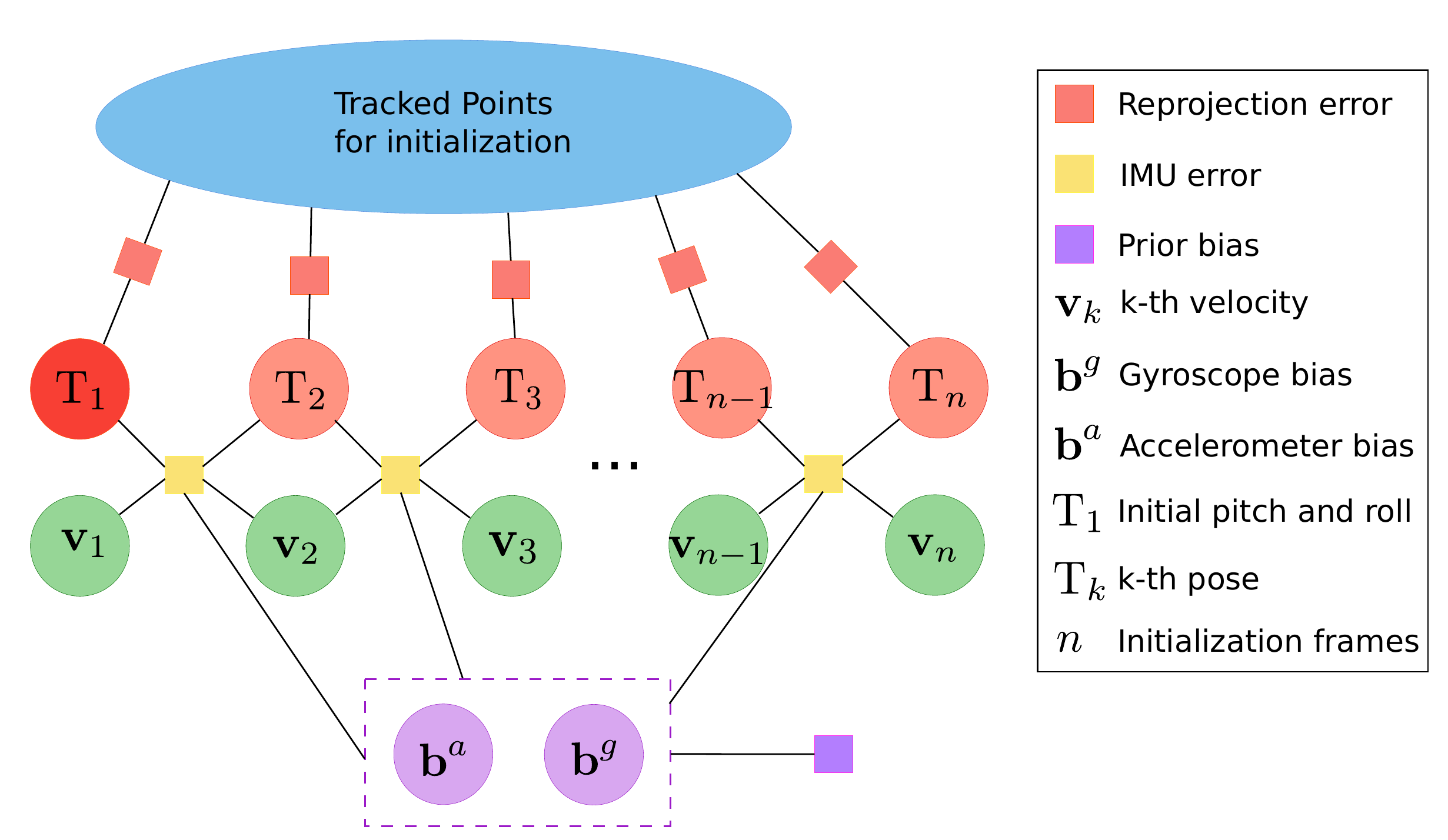}
  \caption{Graph for the first visual-inertial BA. The  body poses and points included in the optimization are the same  used in the initial solution.}
  \label{fig:firstBA}
\end{figure}

After finding the initial parameters ($\textbf{g}$, $\textbf{b}^g$, $\textbf{v}_1$, $\{ \lambda_j^i \}$) we build a visual-inertial BA problem with the same $n$ body poses and $m$ points from the previous step  (see figure \ref{fig:firstBA}). We set the $z$ axis in the estimated gravity direction. All body poses have six optimizable variables $ (\phi,\textbf{t}) \in \mathfrak{se}(3)$ except the first one, which has only two (pitch and roll) since translation and yaw have been fixed in order to remove the four gauge freedoms inherent to the visual inertial problem (initial position and yaw). Body velocities are also included in the optimization task, and they evolve according to the inertial measurements. Initial estimations for each vertex are added using results from the \textit{MK-solution}.
In addition, accelerometer bias $\textbf{b}^a$ is included in this optimization, but similarly to $\textbf{b}^g$ it is assumed to be constant for all frames. Previous $\textbf{b}^g$ estimation from \textit{MK-solution} step is included by means of a prior, as well as $\textbf{b}^a$ is forced to be close to zero.  We call this optimization \textit{first BA} or simply \textit{BA1}. Analytic expression for Jacobians, found in \cite{forster2015imu}, are used for IMU residuals, while Jacobians for the reprojection error have been derived analytically, taking into account that we are optimizing body pose and not camera pose.

Usually this optimization provides a better initialization solution. However, if the motion performed gives low observability of the IMU variables, the optimization can converge to arbitrarily bad solutions. For example this happens in case of pure rotational motion or non-accelerated motions \cite{martinelli2014closed}. In order to detect these failure cases we propose an \textit{observability test}, where we analyze the uncertainty associated to estimated variables. This could be done by analyzing the covariance matrix of the estimated variables and checking if its singular values are small enough. However, this would require to invert the information matrix, i.e. the Hessian matrix from \textit{first BA}, which has high dimensions ($3m+6+9n-4$), being computationally too expensive. Instead, we perform the observability test  imposing a minimal threshold to all singular values of the Hessian matrix associated to our \textit{first BA}. The Hessian can be built from the Jacobian matrices associated to each edge in the graph, as explained next.

\begin{figure}[]
  \centering
  \includegraphics[width=0.48\textwidth]{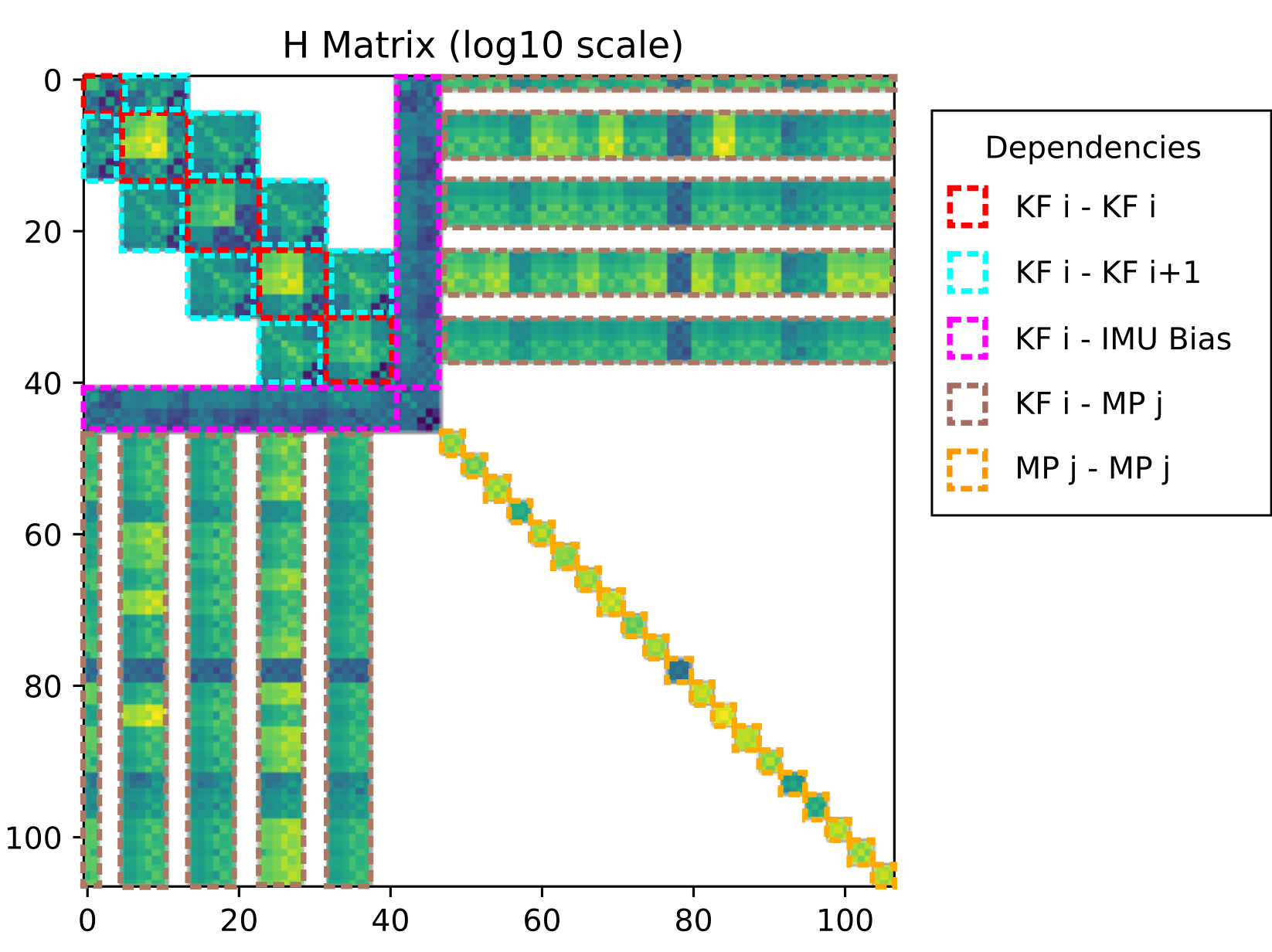}
  \caption{Example of Hessian matrix for an initial map with 5 keyframes (KF) and 20 map points (MP). One can distinguish different blocks, outlined with dashed lines. In the top-left part, we have the diagonal blocks of each keyframe (red), blocks relating consecutive keyframes, due to the IMU measurements (blue), and blocks relating keyframes and IMU biases (pink). In the bottom-right part, there are only the diagonal blocks of the map points (orange). Out-of-diagonal terms relate map points with the keyframes that observe them (brown). In this example all cameras observe all features.}
  \label{fig:Hessian}
\end{figure}

Denote $\{\textbf{x}_1 \dots \textbf{x}_p\}$ the set of $p$ states, and $\{\textbf{e}_1 \dots \textbf{e}_q\}$ the set of $q$ measurements which appear in the \textit{first BA}. Let's call $\mathcal{E}_i$ the set of measurement where state $i$ is involved. The Hessian block matrix for states $i$ and $j$, taking a first order approximation, can be built as follows:

\begin{equation}
\textbf{H}_{i,j} \approx \sum_{\textbf{e} \in \mathcal{E}_i \cap \mathcal{E}_j} \textbf{J}_{i,\textbf{e}}^T \Omega_{\textbf{e}} \textbf{J}_{j,\textbf{e}}
\end{equation}
 
\noindent where $\Omega_{\textbf{e}}$ stands for the information matrix of the $\textbf{e}$ measurement, and $\textbf{J}_{i,\textbf{e}}$ for the Jacobian of the $\textbf{e}$ measurement w.r.t. $i$-th state. In order to have a non-zero $(i,j)$ block matrix, there must to be an edge between $i$ and $j$ node in the graph (measurement depending on both variables) as shown in figure \ref{fig:Hessian}.
 
\begin{figure}[]
  \centering
  \includegraphics[width=0.52\textwidth]{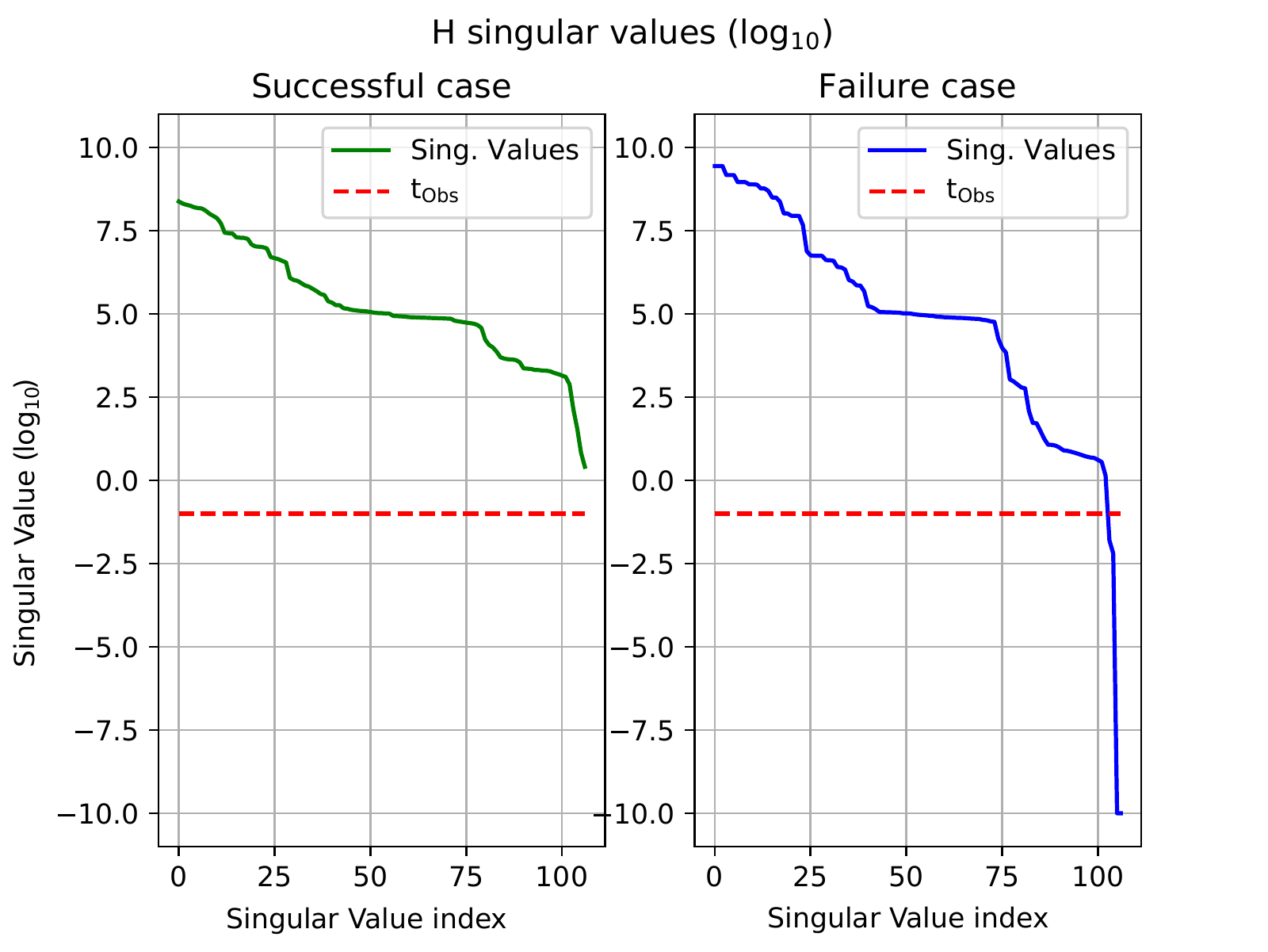}
  \caption{Singular values of the information matrix for a successful initialization and a failure case on the EuRoC V103 sequence. The successful case has a RMSE ATE error of 3.16 $\%$ in the initialization trajectory, and corresponds to a translation and rotation motion. The failure case has an error of 64.99 $\%$ and corresponds to an almost pure rotational motion. We draw the observability threshold used $t_{obs} = 0.1$}
  \label{fig:SingularValues}
\end{figure}

Applying the SVD decomposition to $\textbf{H}$ and looking at the smallest singular value one can determine if the performed motion guarantees observability of all the IMU variables. Hence, we discard all initializations where the smallest Hessian singular value falls below a threshold denoted by $t_{obs}$. If this \textit{observability test} is not passed, we discard the initialization attempt. Examples of a successful and a rejected case are shown in figure \ref{fig:SingularValues}.

 \subsection{Consensus test and second BA}
  \label{sec:BA2_Cons}
As we have noted before, not all $M$ tracked features have been used in \textit{MK-solution} and \textit{first BA} steps, but only $m$ features. To take advantage of these extra unused tracked points, we propose to perform a \textit{consensus test} in order to detect initializations which have been performed using spurious data, such as bad tracked features.

First, the 3D point position of each unused track is triangulated between the two most distant frames which saw the point, by mean of Least-Squares triangulation using a SVD decomposition \cite{szeliski2010computer}. Only tracks with parallax greater than 0.01 radians are used. Then we re-project each 3D point into all the frames which observe it, compute the residual re-projection error, and perform  a $\chi^2(95\%)$ test with  $2n_i-3$ degrees of freedom, where $n_i$ is the number of frames which observe this point. The \textit{consensus test} is performed counting the percentage of inliers: if it is bigger than a threshold $t_{cons}$ we consider that the proposed solution is accurate, if not, we discard the initialization attempt.

If the \textit{consensus test} is successful, we perform a \textit{second BA} (or simply \textit{BA2})  including the $m$ points used in the initial solution plus all the points which have been triangulated and detected as inliers, having a total of $M'$ points. The graph for this optimization is similarly built than in case of \textit{BA1} but with more points.

\subsection{Map initialization}
\label{sec:MapInit}

After this second BA, the keyframe poses are accurate enough, but we only have a few points to initialize the map. Before launching the whole ORB-SLAM visual-inertial system, we triangulate new points aiming to densify the point cloud and to ease the posterior tracking operation. Since we already have the keyframe poses, we extract ORB features in each keyframe and perform an epipolar search in each other, using the ORB descriptor. All these new points, together with the $M'$ points from \textit{BA2}, are promoted to map points, and the $n$ frames used for initialization are promoted to map keyframes. The covisiblity graph \cite{mur2015orb} of this new map is also created, taking into account the observations of points.

\section{Experiments}

The most important parameters of our method are shown in table \ref{tab:Implementation}. Our implementation uses ORB-SLAM visual-inertial \cite{mur2017visual} with its three threads for tracking, mapping and loop closing. Initialization is performed in a parallel thread, thus it has no effect in the real time tracking thread. For \textit{MK-solution} we use Eigen C++ library, while for graph optimization of \textit{BA1} and \textit{BA2} we use g2o C++ library \cite{kummerle2011g}. Experiments have been run in V1 dataset from EuRoC \cite{burri2016euroc} using a Intel Core i7-7700 computer with 32 GB of memory.

%\begingroup
\begin{table}
\centering
\caption {\label{tab:Implementation} Parameters of our initialization algorithm}
\begin{tabular}{lcc}
\hline
 Total number of tracks   & $M$ & 200 \\
 \textit{Track-length test} (in pixels)   & $l$ & 200 \\
 Tracks used for \textit{MK-solution}   & $m$ & 20 \\
 Keyframes used for \textit{MK-solution}  & $n$ & 5 \\
 \textit{Observability test}: Singular value threshold    & $t_{obs}$ & 0.1 \\
\textit{Consensus test}: Inlier threshold   & $t_{cons}$ & 90\% \\
 \hline
 \end{tabular}
\end{table}
%\endgroup

\subsection{Results}
EuRoC dataset provides stereo images and synchronized IMU measures for three different indoor environments, with different complexity. We have tested our method for environment V1 from EuRoC at three difficulty levels. We run two different experiments. 

\begin{figure}
  \centering
  \includegraphics[width=0.52\textwidth]{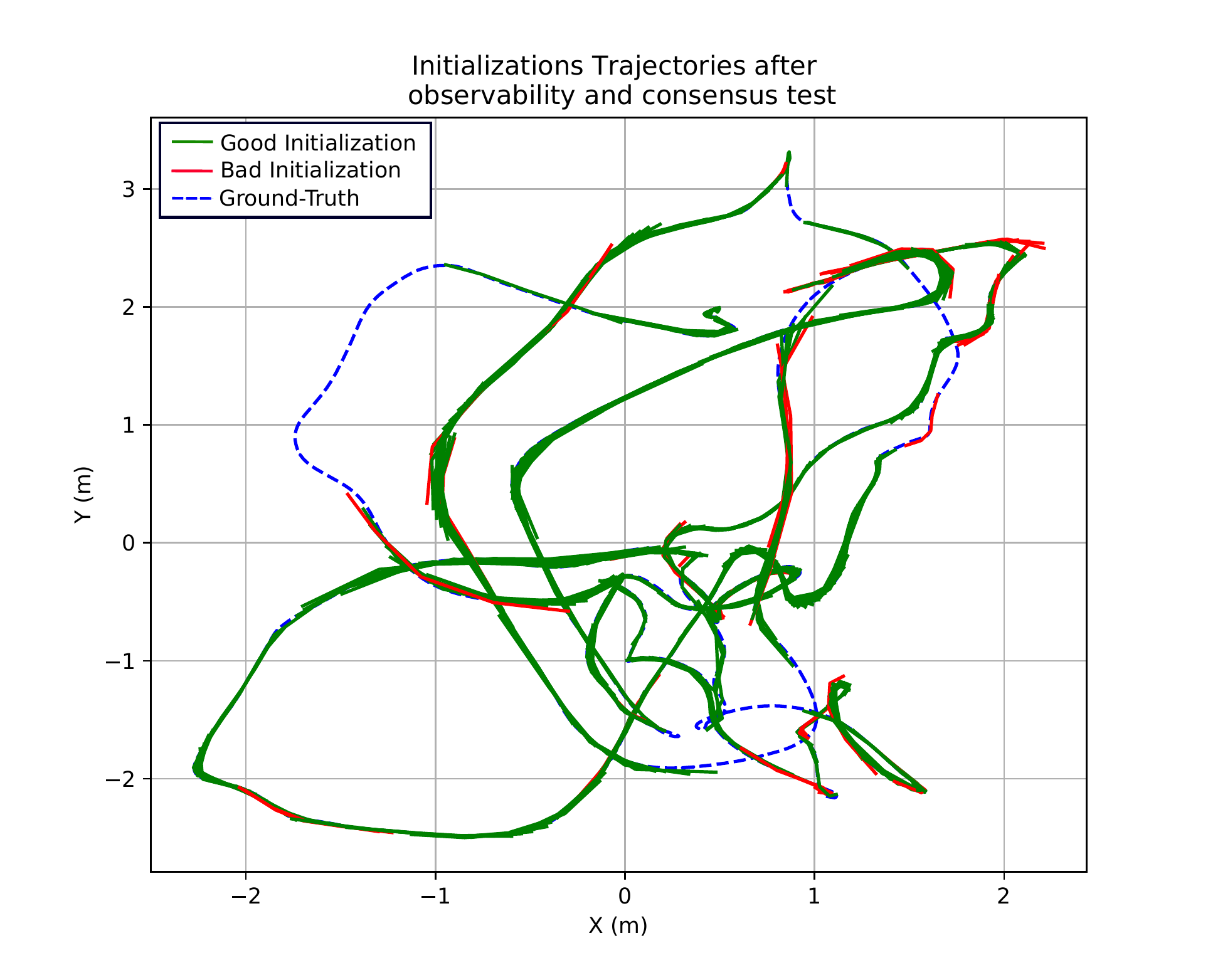}
  \caption{Initializations found along the  EuRoC V101 trajectory, after the \textit{observability and consensus tests}. In blue, ground truth trajectory; in green, estimated initialization trajectories whose RMSE ATE error is lower than 5$\%$; in red, those with a bigger error. Our method was able to find 511 correct initializations along the whole trajectory, running in real time.}
  \label{fig:trajectory}
\end{figure}

In a first experiment, we try to initialize as often as possible in real time. Along the whole trajectory, every time the tracking thread has $m$ tracks with length $l$, if the initialization thread is idle, a new initialization attempt is launched. Figure \ref{fig:trajectory} shows the  initializations found for trajectory V101 after the  \textit{observability and consensus test}. We show in red trajectories which have a RMSE ATE \cite{sturm2012benchmark} error bigger than 5\% of the initialization trajectory length. We can see in the figure that our initialization algorithm is successful almost along all the trajectory. The parts without initializations are due to rejection from observability or consensus test. 

In table \ref{tab:tableRes1} we show the main numerical results of these experiments with the three V1 sequences. RMSE ATE \cite{sturm2012benchmark} is expressed in percentage over the length of the initialization trajectory. Below each sequence name we show successful initializations over the total number. First thing to notice is the large number of initialization attempts. For example, in sequence V101 which lasts 130 seconds, up to 728 initializations are computed, and 511 of them have passed the  \textit{observability and consensus test}. The table shows that the original Martinelli-Kaiser solution obtains average scale errors between 32.9\% and 156.7\% on these sequences. This error can be reduced until  8.8\% to 24.5\% applying the two rounds of visual-inertial BA proposed here. More interestingly, applying the novel \textit{observability and consensus tests}, inaccurate  initializations are consistently rejected, and the average scale error is reduced to around 5\% for all sequences, a very significant improvement over the original method. The ATE error is also drastically reduced after both tests. 

Considering the initialization time we see an evident difference between V101, that requires initialization trajectories of 2.2 seconds in average, and V102 and V103 where 1 second is enough. In these two last sequences motion is faster and the \textit{track-length test} is satisfied in less time than in the first sequence, where the quad-copter is flying at low speed.

Regarding the computational cost, the average CPU required to solve the initialization is less than 85ms for sequences V102 and V103, and around 121ms for V101, due to the longer preintegration period. In all cases, the \textit{MK-solution} step takes around 75\% of the total initialization CPU time.

\begingroup
\begin{table}
\scriptsize
\centering
\caption {\label{tab:tableRes1} Results of exhaustive initialization tests over the three V1 EuRoC sequences. }
\setlength\tabcolsep{1.5pt}
\begin{tabular}{ccrrrrrrr}
  \multicolumn{9}{c}{V1 EuRoC Dataset}  \\ \hline
 {} & {} & \multicolumn{2}{c}{ \begin{tabular}{@{}c@{}} After \\ \textit{track-length test} \end{tabular}}  & & \multicolumn{2}{c}{\begin{tabular}{@{}c@{}} After \textit{Observ} \\ + \textit{Cons. test} \end{tabular}}    \\ 
\cline{3-4} \cline{6-7} 
       \begin{tabular}{@{}c@{}} Seq. \\ Name \end{tabular} & &  \begin{tabular}{@{}c@{}}RMSE \\ ATE (\%) \end{tabular}  & \begin{tabular}{@{}c@{}}Scale \\ error (\%) \end{tabular} & & \begin{tabular}{@{}c@{}}RMSE \\ ATE (\%) \end{tabular} & \begin{tabular}{@{}c@{}}Scale \\ error (\%) \end{tabular} & \begin{tabular}{@{}c@{}}CPU \\ time (ms) \end{tabular} & \begin{tabular}{@{}c@{}}Trajectory \\ time (s) \end{tabular}   \\    \hline 
        
 {} & \textit{MK-solution}   & 9.176  & 32.998 & & 7.749 & 25.104  & 95.082 & 2.235    \\
  \begin{tabular}{@{}c@{}} V101 \\ (511/728)\end{tabular}  &  \begin{tabular}{@{}c@{}}\textit{MK-solution} \\ + \textit{BA1} \end{tabular}  & 3.977 & 10.719 & & 2.352 & 6.471 & 104.114 & 2.235      \\
  {} & \begin{tabular}{@{}c@{}}\textit{MK-solution} \\ + \textit{BA1\&2} \end{tabular}  & 3.270 & 8.816 & & \textbf{2.036} & \textbf{5.496} & 120.983 & 2.235      \\ \hline %\hline

  {} & \textit{MK-solution}   & 12.025 & 156.751 & & 6.760 & 48.926 & 60.285 & 0.968    \\
 \begin{tabular}{@{}c@{}} V102 \\ (101/395)\end{tabular} & \begin{tabular}{@{}c@{}}\textit{MK-solution} \\ + \textit{BA1} \end{tabular}   & 6.338 & 25.252 & & 2.541 & 7.195 & 70.963 & 0.968      \\
 {} & \begin{tabular}{@{}c@{}}\textit{MK-solution} \\ + \textit{BA1\&2} \end{tabular}   & 5.149 &  20.341 &  & \textbf{1.935} & \textbf{5.497} & 84.443 & 0.968      \\ \hline %\hline
        
  {} & \textit{MK-solution}   & 47.928 & 128.008 & & 6.634 & 21.691 & 62.160 & 1.070    \\
   \begin{tabular}{@{}c@{}} V103 \\ (71/336)\end{tabular}  & \begin{tabular}{@{}c@{}}\textit{MK-solution} \\ + \textit{BA1} \end{tabular}    & 71.774 & 28.160 & & 2.475 & 6.836 & 73.301 & 1.070      \\
 {} & \begin{tabular}{@{}c@{}}\textit{MK-solution} \\ + \textit{BA1\&2} \end{tabular}    & 71.068 & 24.556 & & \textbf{1.870} & \textbf{5.259} & 84.676 & 1.070      \\ \hline %\hline

\end{tabular}
\end{table}
\endgroup

%\begingroup
%\begin{table}
%\footnotesize
%\centering
%\caption {\label{tab:tableRes2} Results of VI-SLAM using our initialization. Each dataset has been run five times and errors are for frame trajectory. Mean values are displayed. }
%\setlength\tabcolsep{1.5pt}
%\begin{tabular}{crrrrrrrrrrr}
%  \multicolumn{12}{c}{EuRoC Dataset}  \\ \hline
%  & \multicolumn{2}{c}{ \begin{tabular}{@{}c@{}} After \\ initialization \end{tabular}}  & & \multicolumn{2}{c}{After BA 5s} & & \multicolumn{2}{c}{After BA 10s} & & \multicolumn{2}{c}{After BA 15s} \\ 
%\cline{2-3} \cline{5-6} \cline{8-9} \cline{11-12} 
%       \begin{tabular}{@{}c@{}} Seq. \\ Name \end{tabular} & \begin{tabular}{@{}c@{}}RMSE \\ ATE \\ (m) \end{tabular}  & \begin{tabular}{@{}c@{}}Scale \\ error \\ (\%) \end{tabular} & & \begin{tabular}{@{}c@{}}RMSE \\ ATE \\ (m) \end{tabular}  & \begin{tabular}{@{}c@{}}Scale \\ error \\ (\%) \end{tabular} & & \begin{tabular}{@{}c@{}}RMSE \\ ATE \\ (m) \end{tabular}  & \begin{tabular}{@{}c@{}}Scale \\ error \\ (\%) \end{tabular} & & \begin{tabular}{@{}c@{}}RMSE \\ ATE \\ (m) \end{tabular}  & \begin{tabular}{@{}c@{}}Scale \\ error \\ (\%) \end{tabular} \\    \hline 
%V1\_01\_easy & 0.0183 & 4.99 & & 0.0200 & 1.85 & & 0.0170 &	0.84 & & \textbf{0.0232} & \textbf{1.23}
%\\
%V1\_02\_medium & 0.0364 & 7.38 & & 0.0076 & 3.67 & & 0.0162 & 0.71 & & \textbf{0.0146} & \textbf{0.24} \\
%V1\_03\_difficult & 0.0043 & 4.80 & & 0.0129 & 2.50 & & 0.0120 & 0.27 & & \textbf{0.0278} & \textbf{1.07} \\ 
%\end{tabular}
%\end{table}
%\endgroup

In table \ref{tab:tableTCPU} we show computational times for our method which uses preintegration with first order bias correction from \cite{forster2015imu}. Compared with using the original formulation from Martinelli and Kaiser, computing time is reduced by 60\%.

In a second experiment, we launch visual-inertial ORB-SLAM \cite{mur2017visual} and we retrieve the RMSE ATE and the scale error just after the proposed initialization, and after performing full visual-inertial BA at 5 seconds and 10 seconds from the first keyframe timestamp. We can see in table \ref{tab:tableRes2} that all three sequences converge to scale error smaller than 1$\%$ after 10 seconds, confirming that our initialization method is accurate enough to launch visual-inertial SLAM. An example of Visual-Inertial ORBSLAM \cite{mur2017visual} using our proposed initialization can be found in the accompanying video.

Compared with the initialization method proposed in \cite{mur2017visual}, our method requires trajectories of 1 or 2 seconds instead of 15 seconds, uses less CPU time, and is able to successfully initialize in sequence V103, where the previous method failed.

\begingroup
\begin{table}
\centering
\caption {\label{tab:tableTCPU} Comparison of running time for \textit{MK Solution+BA1+BA2} repeating IMU integration in each iteration and using preintegration with first order bias correction \cite{forster2015imu}.}
\begin{tabular}{cccc}
  \multicolumn{4}{c}{V101 EuRoC Dataset}  \\ \hline
   & \multicolumn{3}{c}{CPU time (ms)}     \\  
  \cline{2-4} & Mean & Std & Max \\ \hline
 Reintegrating each time  & 301.302 & 91.974 & 678.886 \\
 Using first order correction  & 120.983 & 27.609 & 214.989     \\ \hline
 \\
 %\hline
\end{tabular}
\end{table}
\endgroup

\begingroup
\begin{table}
\centering
\caption {\label{tab:tableRes2} Results of VI-SLAM using our initialization (average errors on five executions are shown).}
\setlength\tabcolsep{3pt}
\begin{tabular}{crrrrrrrr}
  \multicolumn{9}{c}{V1 EuRoC Dataset}  \\ \hline
  & \multicolumn{2}{c}{ \begin{tabular}{@{}c@{}} After \\ initialization \end{tabular}}  & & \multicolumn{2}{c}{After BA 5s} & & \multicolumn{2}{c}{After BA 10s} \\ 
\cline{2-3} \cline{5-6} \cline{8-9} 
       \begin{tabular}{@{}c@{}} Seq. \\ Name \end{tabular} & \begin{tabular}{@{}c@{}}RMSE \\ ATE \\ (m) \end{tabular}  & \begin{tabular}{@{}c@{}}Scale \\ error \\ (\%) \end{tabular} & & \begin{tabular}{@{}c@{}}RMSE \\ ATE \\ (m) \end{tabular}  & \begin{tabular}{@{}c@{}}Scale \\ error \\ (\%) \end{tabular} & & \begin{tabular}{@{}c@{}}RMSE \\ ATE \\ (m) \end{tabular}  & \begin{tabular}{@{}c@{}}Scale \\ error \\ (\%) \end{tabular} \\    \hline 
V1\_01\_easy & 0.0183 & 4.99 & & 0.0200 & 1.85 & & \textbf{0.0170} &	\textbf{0.84} \\
V1\_02\_medium & 0.0364 & 7.38 & & 0.0076 & 3.67 & & \textbf{0.0162} & \textbf{0.71} \\
V1\_03\_difficult & 0.0043 & 4.80 & & 0.0129 & 2.50 & & \textbf{0.0120} & \textbf{0.27} \\ 
 \hline %\hline
\end{tabular}
\end{table}
\endgroup

\section{Conclusions}
We have proposed a fast joint monocular-inertial initialization method, based on the work of Martinelli \cite{martinelli2014closed} and Kaiser et al. \cite{kaiser2017simultaneous}. We have adapted it to be more general, allowing incomplete feature tracks, and more computationally efficient using the IMU preintegration method of Forster et al. \cite{forster2015imu}. Our results show that the original Martinelli-Kaiser technique does not provide a good enough initialization in most practical scenarios, hence we have proposed two visual-inertial BA steps to improve the solution and two novel tests to detect bad initializations. These techniques have proven to be worth it, reducing scale error down to 5$\%$ and rejecting bad initializations. Solutions found after those steps is good enough to launch Visual-Inertial ORBSLAM \cite{mur2017visual} and converge to very accurate maps.

In summary, we have developed a fast method for joint initialization of monocular-inertial SLAM, using trajectories of 1 to 2 seconds, that is much more accurate and robust than the original technique  \cite{kaiser2017simultaneous}, with a maximum computational cost of 215ms.

As future work we would like to investigate the adaptation of the initialization method to the stereo case, taking into account that scale is directly observable from the images. We are also interested in taking profit of gyroscope readings for tracking, even before the initialization has been performed. Finally, we would like to test the initialization performance in in more difficult scenarios with our own acquired sequences.
                                                                 \addtolength{\textheight}{-15cm}   % This command serves to balance the column lengths

\bibliographystyle{IEEEtran}
\bibliography{VI_initialization,IEEEabrv}

\begin{thebibliography}{10}
\providecommand{\url}[1]{#1}
\csname url@rmstyle\endcsname
\providecommand{\newblock}{\relax}
\providecommand{\bibinfo}[2]{#2}
\providecommand\BIBentrySTDinterwordspacing{\spaceskip=0pt\relax}
\providecommand\BIBentryALTinterwordstretchfactor{4}
\providecommand\BIBentryALTinterwordspacing{\spaceskip=\fontdimen2\font plus
\BIBentryALTinterwordstretchfactor\fontdimen3\font minus
  \fontdimen4\font\relax}
\providecommand\BIBforeignlanguage[2]{{%
\expandafter\ifx\csname l@#1\endcsname\relax
\typeout{** WARNING: IEEEtran.bst: No hyphenation pattern has been}%
\typeout{** loaded for the language `#1'. Using the pattern for}%
\typeout{** the default language instead.}%
\else
\language=\csname l@#1\endcsname
\fi
#2}}

\bibitem{martinelli2014closed}
A.~Martinelli, ``Closed-form solution of visual-inertial structure from
  motion,'' \emph{International Journal of Computer Vision}, vol. 106, no.~2,
  pp. 138--152, 2014.

\bibitem{kaiser2017simultaneous}
J.~Kaiser, A.~Martinelli, F.~Fontana, and D.~Scaramuzza, ``Simultaneous state
  initialization and gyroscope bias calibration in visual inertial aided
  navigation,'' \emph{IEEE Robotics and Automation Letters}, vol.~2, no.~1, pp.
  18--25, 2017.

\bibitem{li2013high}
M.~Li and A.~I. Mourikis, ``High-precision, consistent {EKF}-based
  visual-inertial odometry,'' \emph{The International Journal of Robotics
  Research}, vol.~32, no.~6, pp. 690--711, 2013.

\bibitem{leutenegger2015keyframe}
S.~Leutenegger, S.~Lynen, M.~Bosse, R.~Siegwart, and P.~Furgale,
  ``Keyframe-based visual--inertial odometry using nonlinear optimization,''
  \emph{The International Journal of Robotics Research}, vol.~34, no.~3, pp.
  314--334, 2015.

\bibitem{mur2017visual}
R.~Mur-Artal and J.~D. Tard{\'o}s, ``Visual-inertial monocular {SLAM} with map
  reuse,'' \emph{IEEE Robotics and Automation Letters}, vol.~2, no.~2, pp.
  796--803, 2017.

\bibitem{qin2017robust}
T.~Qin and S.~Shen, ``Robust initialization of monocular visual-inertial
  estimation on aerial robots,'' in \emph{IEEE/RSJ International Conference on
  Intelligent Robots and Systems (IROS)}, 2017, pp. 4225--4232.

\bibitem{lupton2012visual}
T.~Lupton and S.~Sukkarieh, ``Visual-inertial-aided navigation for high-dynamic
  motion in built environments without initial conditions,'' \emph{IEEE
  Transactions on Robotics}, vol.~28, no.~1, pp. 61--76, 2012.

\bibitem{forster2015imu}
C.~Forster, L.~Carlone, F.~Dellaert, and D.~Scaramuzza, ``{IMU} preintegration
  on manifold for efficient visual-inertial maximum-a-posteriori estimation,''
  in \emph{Robotics: Science and Systems}, 2015.

\bibitem{burri2016euroc}
M.~Burri, J.~Nikolic, P.~Gohl, T.~Schneider, J.~Rehder, S.~Omari, M.~W.
  Achtelik, and R.~Siegwart, ``The {EuRoC} micro aerial vehicle datasets,''
  \emph{The International Journal of Robotics Research}, vol.~35, no.~10, pp.
  1157--1163, 2016.

\bibitem{rublee2011orb}
E.~Rublee, V.~Rabaud, K.~Konolige, and G.~Bradski, ``{ORB}: An efficient
  alternative to {SIFT} or {SURF},'' in \emph{IEEE International Conference on
  Computer Vision (ICCV)}, 2011, pp. 2564--2571.

\bibitem{lucas1981iterative}
B.~D. Lucas, T.~Kanade, \emph{et~al.}, ``An iterative image registration
  technique with an application to stereo vision,'' in \emph{Int. Joint. Conf.
  on Artificial Intelligence (IJCAI)}, 1981, pp. 674--679.

\bibitem{szeliski2010computer}
R.~Szeliski, \emph{Computer vision: algorithms and applications}.\hskip 1em
  plus 0.5em minus 0.4em\relax Springer Verlag, London, 2011.

\bibitem{mur2015orb}
R.~Mur-Artal, J.~Montiel, and J.~D. Tard{\'o}s, ``{ORB-SLAM}: a versatile and
  accurate monocular {SLAM} system,'' \emph{IEEE Transactions on Robotics},
  vol.~31, no.~5, pp. 1147--1163, 2015.

\bibitem{kummerle2011g}
R.~K{\"u}mmerle, G.~Grisetti, H.~Strasdat, K.~Konolige, and W.~Burgard, ``g2o:
  A general framework for graph optimization,'' in \emph{IEEE International
  Conference on Robotics and Automation (ICRA)}, 2011, pp. 3607--3613.

\bibitem{sturm2012benchmark}
J.~Sturm, N.~Engelhard, F.~Endres, W.~Burgard, and D.~Cremers, ``A benchmark
  for the evaluation of {RGB-D} {SLAM} systems,'' in \emph{IEEE/RSJ
  International Conference on Intelligent Robots and Systems (IROS)}, 2012, pp.
  573--580.

\end{thebibliography}

\end{document}